\pdfoutput=1

\documentclass[11pt]{article}

\usepackage{ACL2023}

\usepackage{times}
\usepackage{latexsym}

\usepackage[T1]{fontenc}

\usepackage[utf8]{inputenc}

\usepackage{microtype}

\usepackage{inconsolata}

\usepackage[utf8]{inputenc}
\usepackage[T1]{fontenc}
\usepackage{hyperref}
\usepackage{booktabs}
\usepackage{amsfonts}
\usepackage{nicefrac}
\usepackage{microtype}
\usepackage{xcolor}
\usepackage{amsthm}
\usepackage{amsmath}
\usepackage{float}
\usepackage{graphicx}
\usepackage{grffile}
\usepackage{wrapfig}
\usepackage{makecell}
\usepackage{caption}
\usepackage{subcaption}
\usepackage{bm}
\usepackage{multirow}
\usepackage{microtype}
\usepackage{mathtools}
\usepackage{tikz}
\usepackage{gensymb}
\usepackage{bbm}
\usepackage{amssymb}
\usepackage{enumitem}
\usepackage{placeins}
\usepackage{algorithm,algpseudocode}
\usepackage{xspace}
\usepackage{rotating}
\usepackage{adjustbox}
\usepackage{diagbox}
\mathtoolsset{showonlyrefs}

\def\expected{\mathbb{E}}
\renewcommand{\vec}[1]{\boldsymbol{\mathbf{#1}}}
\def\x{\vec{x}}
\def\y{\vec{y}}

\def\r{\vec{r}}
\def\p{\vec{p}}
\def\e{\vec{e}}

\DeclareMathOperator*{\argtopk}{arg-topk}
\def\KL#1#2{\textnormal{KL}\big({#1} | {#2}\big)}

\newcommand{\eqn}[1]{\begin{align}#1\end{align}}

\def\masked{\xi}
\def\*#1{\ensuremath{\mathcal{#1}}}

\def\M{\textnormal{M}}
\def\loss{\mathcal{L}}
\newcommand{\norm}[1]{\lVert#1\rVert}

\title{DiMS: Distilling Multiple Steps of
Iterative Non-Autoregressive Transformers for Machine Translation} 

\author{Sajad Norouzi\thanks{\ \ Equal contribution.} \\
  Layer6 AI\\
  \texttt{sajad@layer6.ai} \\
   \And
   Rasa Hosseinzadeh$^*$ \\
   Layer6 AI \\
   \texttt{rasa@layer6.ai} \\
   \And
   Felipe Pérez \\
   Layer6 AI \\
   \texttt{felipe@layer6.ai} \\
   \And
   Maksims Volkovs \\
   Layer6 AI \\
   \texttt{maks@layer6.ai} \\}

\begin{document}
\maketitle
\begin{abstract}
The computational benefits of iterative non-autoregressive transformers
decrease as the number of decoding steps increases. As a remedy, we introduce
\textbf{Di}still \textbf{M}ultiple \textbf{S}teps (\textbf{DiMS}), a simple yet
effective distillation technique to decrease the number of required steps to
reach a certain translation quality. The distilled model enjoys the
computational benefits of early iterations while preserving the enhancements
from several iterative steps. DiMS relies on two models namely student and
teacher. The student is optimized to predict the output of the teacher after
multiple decoding steps while the teacher follows the student via a slow-moving
average. The moving average keeps the teacher’s knowledge updated and enhances
the quality of the labels provided by the teacher. During inference, the
student is used for translation and no additional computation is added. We
verify the effectiveness of DiMS on various models obtaining 7.8 and 12.9 BLEU
points improvements in single-step translation accuracy on distilled and raw
versions of WMT'14 De-En. Full code for this work is available
here:~\url{https://github.com/layer6ai-labs/DiMS}.

\end{abstract}

\section{Introduction}
\label{sec:intro}

Neural machine translation models typically follow an autoregressive decoding
strategy, generating the target sentence one token at a time. This sequential
nature makes the inference process slow and dependent on the output sequence
length. To address this limitation~\citet{gu2018non} introduces the
Non-Autoregressive Transformer (NAT). NAT generates the entire target sentence
in parallel, reducing the latency by an order of magnitude. NAT can be
considered as a member of a broader family of iterative non-autoregressive
Transformers (iNAT)~\citep{lee2020iterative, stern2019insertion,
ghazvininejad2019mask} where the number of decoding steps is fixed and
independent of the sequence length. By tuning the number of decoding steps, one
can control the trade-off between speed and quality. While iNATs can be
considered as efficient alternatives to their autoregressive counterparts,
\citet{kasai2020deep} shows that autoregressive models can be sped up without
loss in accuracy by combining shallow decoders with deep encoders. This
diminishes the computational advantage of iNATs and challenges their
motivation. The focus of recent work has thus been shifted to design single-step NAT
models~\citep{ghazvininejad2020aligned, qian2021glancing, du2021order}.

\begin{figure}
    \centering
    \includegraphics[width=0.9\linewidth]{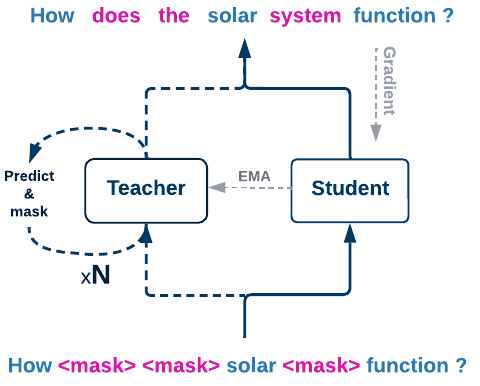}
    \captionof{figure}{DiMS training. The
    student is trained to match the predictions of the teacher after several
    iterative steps. Teacher is updated with an exponential moving average 
    of the student.}
    \label{fig:nesd}
\end{figure}

\begin{figure*}[!h]
    \centering
    \includegraphics[width=0.7\linewidth]{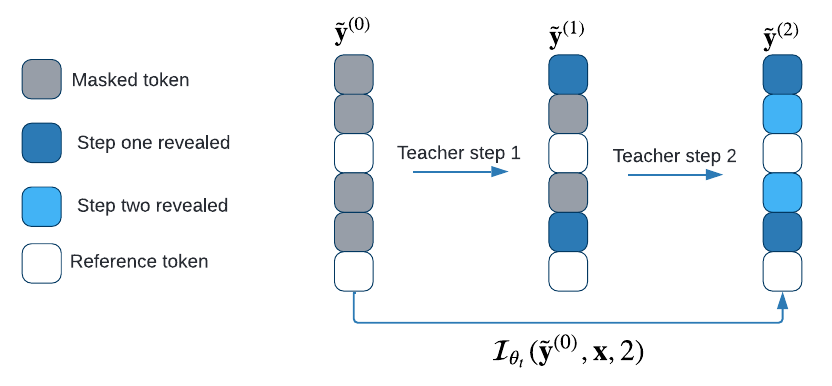}
    \caption{Two iterative steps of the teacher applied to a partially masked sentence.}
    \label{fig:teacher_iterative}

\end{figure*}

In order to preserve the enhancements obtained by multiple decoding iterations
of iNATs, we introduce Distill Multiple Steps (DiMS), a distillation algorithm
applicable to a wide range of iterative models. Given a pre-trained iNAT,
referred to as teacher, a student aims to replicate the behavior of multiple
iterative steps of the teacher with one decoding pass. This process resembles
the well-known knowledge distillation framework\cite{hinton2015distilling}.
However, instead of reducing the number of parameters, we aim to decrease the
number of decoding passes. The final model then enjoys the translation quality
of multi-steps iNAT with the computational efficiency of single-step
translation.

The proposed distillation can be repeated iteratively, where at the end of each
round the newly optimized student becomes the next teacher. While effective,
iterative distillation is slow as it requires multiple rounds of training until
convergence. Alternatively, we propose updating the parameters of the teacher
with an exponential moving average (EMA) of the student. This gradually
transfers the new knowledge learned by the student to the teacher and can be
viewed as a continuous variant of iterative distillation. Figure~\ref{fig:nesd}
depicts the DiMS algorithm.

We demonstrate the effectiveness of our approach on several public datasets by
showing that DiMS obtains substantial improvements on single-step translation
with gains of up to $7.8$ BLEU points on the distilled training dataset, while
the gains on raw datasets are even greater. Notably, we are able to surpass
many leading NAT models designed specifically for single-step translation. We
further show that EMA considerably speeds up training and converges to a
comparable accuracy with iterative distillation in a fraction of epochs.

\section{Background}
\label{sec:background}

In this section, we lay out a formal framework for iNATs. We use the setup
of Conditional Masked Language Models (CMLM). CMLM first introduced
in~\citet{ghazvininejad2019mask} and subsequently adopted in many iNAT
models~\citep{ghazvininejad2020semi, kasai2020non,
huang2021improving}. The source
sentence, target sentence, and target sequence length are denoted by $\x$, $\y$ and $N$,
respectively.

\subsection{Training}

\label{sec:background-train}
Given a partially masked reference sentence $\tilde{\y}$ and the corresponding
source context $\x$, the model is trained to reveal all the masked positions
simultaneously~\citep{ghazvininejad2019mask}. From a probabilistic perspective,
this imposes a conditional independence assumption on the predicted tokens.
Formally, the training loss is:
\eqn{
    \expected_{\tilde{\y} \sim \M(\y)} ~~ {\sum_{i \in \masked(\tilde{\y})}~  -\log p_{\theta} (y_i \vert \x, \tilde{\y}) },
}
where $\M$ is a distribution over all partially masked target sentences and
$\masked$ is a function that returns the set of masked indices. The training
objective above implicitly assumes access to the target sentence length. To
resolve this issue, CMLM trains a parametric model, \emph{length predictor}, to
predict the output length.

\begin{algorithm*}[!h]\caption{DiMS}\label{alg:main}
\begin{algorithmic}
\Require Data set $\mathcal{D}$,
pre-trained model $\phi$,
Hidden state loss factor $\lambda$,
teacher steps $n$,
\\
EMA momentum $\mu$, learning rate $\eta$
\State $\theta_t, \theta_s \leftarrow \phi$ \Comment{Initialize teacher and student}
\While{not converged}
    \State $(\x,\y) \sim \mathcal{D}$ \Comment{Sample data}
    \State $\tilde{\y} \sim \M(\y)$ \Comment{Sample masking}
    \State $\p_{t} \leftarrow \mathcal{I}_{\theta_t}\left(\x, \tilde{\y}, n\right)$ \Comment{Run the teacher for $n$ iterative steps}
    \State $\p_{s} \leftarrow \mathcal{I}_{\theta_s}\left(\x, \tilde{\y}, 1\right)$ \Comment{Run the student for a single step}
    \State  $\loss_{\textrm{DiMS}} \leftarrow  \sum_{i} \KL{\p_{t,i}}{\p_{s, i}}
+ \lambda \norm{\e_{t,i} - \e_{s,i}}^2$
        \Comment{Compute the DiMS loss}
    \State $\theta_s \leftarrow \textrm{Optimizer}(\theta_s, \nabla_{\theta_s} \loss_{\textrm{DiMS}}, \eta)$ 
    \Comment{Gradient based optimization of the student}
    \State $\theta_t \leftarrow (1-\mu)\theta_s + \mu \theta_t$ \Comment{EMA Update of the teacher}
\EndWhile
\end{algorithmic}
\end{algorithm*}

\vspace*{0.2cm}
\subsection{Inference}
\vspace*{0.2cm}
\label{sec:background-inference}

The inference begins by creating a template $\tilde{\y}^{(0)}$ with $\tilde{N}$
masked tokens, where $\tilde{N}$ is the output of the length predictor.
At iteration $t$ of the inference, the model predicts the translation $\r^{(t)}$ given
$\tilde{\y}^{(t-1)}$ and $\x$ as inputs. Depending on the number of decoding
iterations $S$, typically a linear unmasking policy is used where at each step
$\tilde{N}/S$ tokens with the highest probability are revealed. This process is
repeated $S$ times, resulting in a fully revealed sentence.
In other words,
${\tilde{y}^{(t)}_i}$ is set to 
${r^{(t)}_i}$ when
${i \in \argtopk_{k=\frac{\tilde{N}}{S}} 
\left\{ p_{\theta}\left(r_{j}^{(t)} \middle\vert \x, \tilde{\y}^{(t-1)}\right)\right\}}
,$ where $p_\theta$ denotes the output probability of the model.
Otherwise ${\tilde{y}^{(t)}_i}$ stays equal ${\tilde{y}_i^{(t-1)}}.$

Note that multiple length candidates can be considered (e.g.\ $\tilde{N} \pm
1$) with the average token probability as a ranking criterion. This is similar
to beam search in autoregressive models but applied to the output sequence
length. It is referred to as \emph{length beam}.

\section{Distillation of Iterative Non-autoregressive Transformers}
\label{sec:method}

Increasing the number of decoding steps typically improves accuracy, but
diminishes the computational advantage of iNATs. Our objective is to reduce the
number of decoding steps without degrading the performance. More specifically,
we want to condense the translation quality of multiple steps of a teacher into
one decoding pass of a student. For instance, consider an iterative model
(teacher) that uses eight decoding steps. By replicating four steps of the
teacher with one decoding pass, two steps of the student would be sufficient to
reach a similar performance.

The standard way of knowledge distillation would have the teacher generate
soft labels for all intermediate iterations, and optimize the student to
track the teacher's output with fewer steps, but doing such generation on-the-fly
greatly increases the training cost. This process can be moved to a pre-processing
phase, at the cost of large memory requirement. We propose to use partially
masked reference sentences as an approximation to the intermediate predictions
of the teacher, which eliminates the need for several decoding passes or large
memory capacity.

The distillation process starts by initializing the student and the teacher to
the same pre-trained model with parameters $\phi$ i.e.\ ${\theta_s=\theta_t =
\phi}$ where $\theta_s$ and $\theta_t$ denote the parameters of the student and
teacher. Then, the teacher processes a partially masked sentence
$\tilde{\y}$ through $n$ iterative steps with a linear unmasking policy. More
precisely, $i/n$ of the originally masked tokens are revealed up to step $i$ and
after the final pass, no masked token remains. This is similar to the inference
procedure outlined in Section~\ref{sec:background-inference}, but instead of
starting from a fully masked sentence, it starts from a partially masked
one. The student is optimized to match the teacher's soft labels and a
temperature is used to control the smoothness of the labels. With enough
capacity, the student is expected to imitate the behavior of $n$ consecutive
steps of the teacher with one decoding pass.

\begin{figure*}[ht]
\centering
\begin{tabular}{c c c c}
        \includegraphics[width=0.22\linewidth]{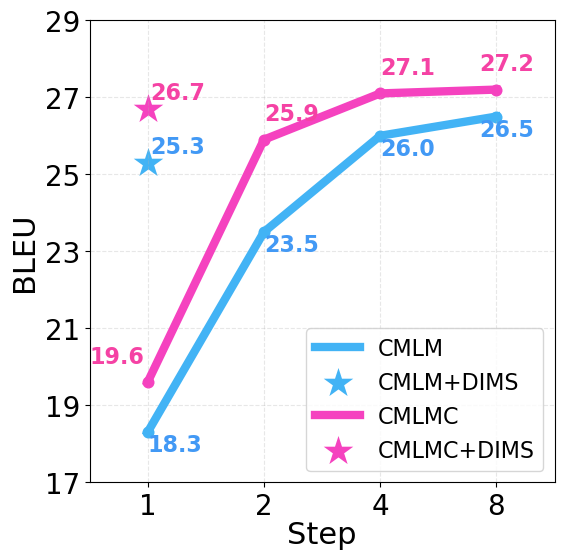} &
        \includegraphics[width=0.22\linewidth]{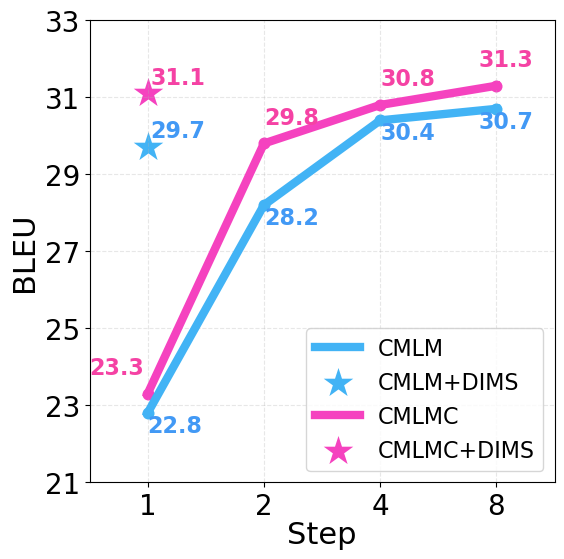} &
        \includegraphics[width=0.22\linewidth]{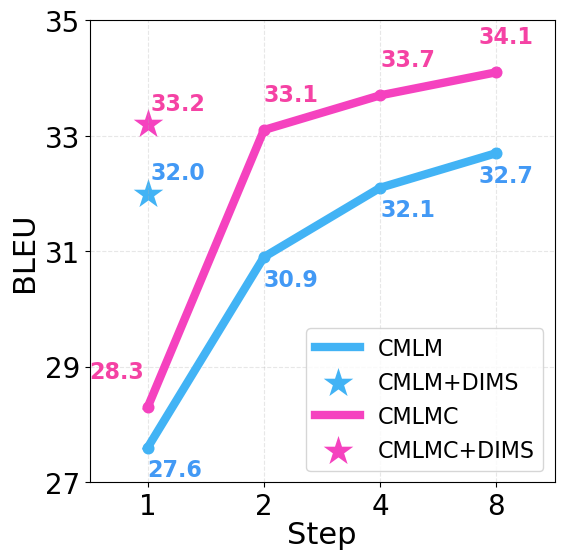} &
        \includegraphics[width=0.22\linewidth]{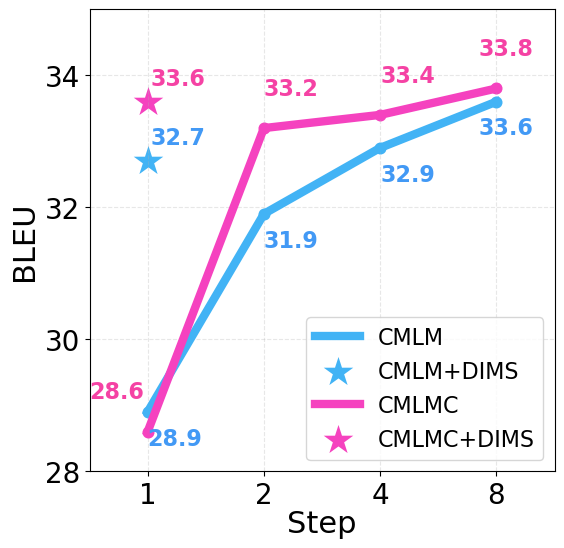} 
        \\
      {\scriptsize a) WMT'14 En-De} & {\scriptsize b) WMT'14 De-En} &
      {\scriptsize c) WMT'16 En-Ro} & {\scriptsize d) WMT'16 Ro-En } 
\end{tabular}
\captionof{figure}{CMLM and CMLMC models distilled with DiMS on four WMT
tasks. For each teacher we plot performance across various decoding steps and
contrast it with a single-step performance of the student.}
\label{fig:all_results}
\end{figure*}

\subsection{Training Loss}
We denote the output distribution after $n$ iterative steps on the partially
masked sentence $\tilde{\y}$ by $ \mathcal{I}_{\theta}\left(\tilde{\y}, \x, n
\right)$ where $\theta$ represents the parameters of the model. The distillation
loss can be described as: $ \sum_{i \in \masked(\tilde{\y})}^{}
\KL{\p_{t,i}}{\p_{s,i}}$ where $\p_t=\mathcal{I}_{\theta_t}\left(\tilde{\y}, \x, n
\right),$ $\p_s=\mathcal{I}_{\theta_s}\left(\tilde{\y}, \x, 1 \right)$ and $i$
in subscript denotes the index in the sentence. Note that the teacher's soft
labels do not come from the same decoding iteration i.e.\ whenever a token is
revealed, the corresponding soft labels are fixed in $p_{t}$. Thus, the student
receives labels from various decoding steps of the teacher.
Figure~\ref{fig:teacher_iterative} depicts the process teacher follows to
produce the labels for two iterative steps. From the student's point of view,
the primary difference between DiMS and CMLM training
(Section~\ref{sec:background-train}) is the use of soft labels generated by the
teacher instead of the ground truth tokens.

To facilitate the distillation, we combine the KL-divergence with the Euclidean
distance of the last layers' hidden states of the teacher and the student. This
transfers the knowledge concealed within the hidden states that might not be
discernible in soft labels. We refer to this as \emph{hidden state loss}.
Similar to the KL-divergence, the hidden state loss is computed over the masked
indices.

To summarize, DiMS training loss has two terms: \textbf{i)} KL-divergence
between distributions predicted by the teacher and the student. \textbf{ii)}
The Euclidean distance between the last hidden states of two models. Denoting
teacher's and student's last hidden state by $\e_{t}$ and $\e_{s}$, DiMS loss
can be written formally as:
\setlength{\textfloatsep}{0.1cm}
\eqn{
    \label{eq:distill-loss}
    &\loss_{\textrm{DiMS}} = \sum_{i} \KL{\p_{t,i}}{\p_{s, i}}
    +\lambda \norm{\e_{t, i} - \e_{s, i}}^2,
}
where
${\p_t = \mathcal{I}_{\theta_t}\left(\tilde{\y}, \x, n \right)}$
and
${\p_s = \mathcal{I}_{\theta_s}\left(\tilde{\y}, \x, 1 \right).}$

The hyper-parameter $\lambda$ controls the contribution of hidden state loss.
When the distillation is completed, the student is used for inference.

\subsection{EMA Update of the Teacher}

\label{sec:method-ema}
As the distillation progresses, the performance gap between multiple steps of
the teacher and a single-pass of the student shrinks, making the teacher's
labels less informative. Two approaches can be considered to sustain the
usefulness of the teacher's labels: \textbf{i)} Increasing the number of
teacher's iterative steps. \textbf{ii)} Restarting the distillation where the
recently optimized student becomes the new teacher and repeating this process
several times, i.e. ${\theta_t^{(n)}\leftarrow \theta_{s}^{(n-1)}}$. The former
makes the training more expensive as the number of sequential steps grows, and
the latter requires repeated distillation rounds leading to a longer training
time.

Instead, we propose updating the teacher with the student's recently learned
knowledge. As the student's single-step output approaches the
teacher's multi-step, the student's multi-step performance would improve as
well, and it is beneficial to use the improved
student as the new teacher. However, replacing the teacher directly with the
student would hurt the training stability, and can lead to a pathological
solution of mapping everything to a constant vector. This degenerate solution
shortcuts the $\loss_{\textrm{DiMS}}$ loss by setting it to a global minimum of
zero. To alleviate this, we update the teacher with a slow-exponential-moving
average of the student, which transfers the new knowledge learned by the
student to the teacher in a controlled manner. The updated teacher now provides
a better training target for the student, creating a positive feedback loop
between the two models. The teacher also benefits from the ensembling effects of the
EMA~\citep{izmailov2018averaging}. Algorithm~\ref{alg:main} outlines the steps
for DiMS training with EMA.

\begin{table*}
\centering
\begin{tabular}{l l c c c c}
\toprule
& \textbf{Model} & \multicolumn{2}{c}{\textbf{WMT'14}}  & \multicolumn{2}{c}{\textbf{WMT'16}}  \\
 & & En-De & De-En & En-Ro & Ro-En  \\
 \toprule
\multirow{9}{*}{\rotatebox[origin=c]{90}{\scriptsize XE-Based}} 
& \footnotesize CMLM \scriptsize \citep{ghazvininejad2019mask} & 18.1  & 21.8  & 27.3 & 28.2  \\
& \footnotesize SMART \scriptsize \citep{ghazvininejad2020semi}  & 18.6 & 23.8 & - & -  \\
& \footnotesize CMLMC \scriptsize \citep{huang2021improving}   & 19.6 & 23.6 & 28.2 & 29.0  \\
& \footnotesize Aux. Reg. \scriptsize \citep{wang2019non}  &  20.7  & 24.8  & - & - \\
& \footnotesize Bag-of-ngram \scriptsize \citep{shao2020minimizing}  & 20.9 & 24.6 & 28.3 & 29.3 \\
& \footnotesize Hint-based Loss \scriptsize \citep{shao2020minimizing} & 21.1 & 25.2 & - & - \\
& \footnotesize Bigram CRF \scriptsize \citep{sun2019fast} & 23.4 & 27.2 & - & - \\
& \footnotesize EM+ODD \scriptsize \citep{sun2020approach} & 24.5 & 27.9 & - & - \\
& \footnotesize ENGINE \scriptsize \citep{tu2020engine} & - & 28.1 & - & 28.2 \\
& \footnotesize GLAT \scriptsize \citep{alignart} & 25.2 & 29.8 & 31.2 & 32.04 \\
  \midrule
& \footnotesize CMLMC + DiMS & 26.7  & 31.1  & 33.2 & 33.6  \\
\midrule
\multirow{5}{*}{\rotatebox[origin=c]{90}{\scriptsize Alignment-Based}}
& \footnotesize Imputer \scriptsize \citep{Saharia2020NonAutoregressiveMT} & 25.8 & 28.4 & 32.3 & 31.7 \\
& \footnotesize AXE \scriptsize \citep{ghazvininejad2020aligned} & 23.5 & 27.9 & 30.8 & 31.5 \\
& \footnotesize OAXE \scriptsize \citep{du2021order} & 26.1 & 30.2 & 32.4 & 33.3 \\
& \footnotesize AlignNART \scriptsize \citep{alignart} & 26.4 & 30.4 & 32.5 & 33.1 \\

& \footnotesize FullyNAT + CTC + GLAT\scriptsize \citep{gu2020fully} & 27.2 & 31.4 & 33.7 & 34.2 \\
& \footnotesize DAT \scriptsize \citep{huang2022directed} & 27.3 & 31.3 & - & -\\
\bottomrule
\end{tabular}
\captionof{table}{Comparison of the single-step test set BLEU score with
  previously published works.}
\label{tab:main}
\end{table*}

\section{Experiments}
\label{sec:expr}
\subsection{Experimental Setup}

We use Fairseq~\citep{ott2019fairseq} for all the experiments and
follow the default data splits. All models are Transformers with
encoder-decoder architecture, each having 6 layers and 512-dimensional hidden
states. Adam optimizer with inverse squared root learning rate scheduler is
used along with mixed precision. EMA and hidden state loss are leveraged with
two iterative steps of the teacher unless otherwise stated. We use early
stopping based on single-step BLEU score on the validation set. The final model
is the average of 5 best checkpoints. Dropout is disabled for the teacher and
the student since empirical improvements are observed. We conduct experiments
on both the raw and distilled dataset that is obtained from an autoregressive
model~\citep{gu2018non}. Training is done with 4 Tesla V100 GPUs (32 GB) and we
report all the hyper-parameters in Section~\ref{sec:sup-hyp} of the appendix.
The extra computational cost of distillation is a small fraction of original
training. We report a detailed comparison in Section~\ref{sec:sup-comp} of the
appendix.

\subsection{Main Results}

Our main experiments are conducted on WMT'14 En-De and WMT'16 En-Ro datasets
with two models: \textbf{i)} CMLM, a pivotal work in iNAT literature showing
the effectiveness of conditional masked language models. \textbf{ii)} CMLMC, a
recent work improving CMLM by incorporating a correction mechanism.
The corresponding official repositories are used to train the teachers. Both models
exploit a length predictor that is conditioned on the encoder's hidden states.
For CMLMC models we use encoder side masking and prediction
\citep{guo2020jointly} to further boost the performance of the teacher. To make
the length predictor compatible with changes in the encoder, we keep the length
predictor loss during distillation.

Figure~\ref{fig:all_results} contrasts the single-step BLEU score of students
with teachers evaluated for various number of decoding steps. DiMS considerably
improves the translation quality of the single-step inference, reducing or
eliminating the gap with multi-step inference. For example, on the WMT'14 De-En
dataset, the single-step of CMLMC+DiMS surpasses the teacher's 4-step
performance. We compared our best single-step model with strong baselines
in Table~\ref{tab:main} showing the effectiveness of our approach. DiMS outperforms all
cross-entropy based models and makes cross-entropy based models competitive with 
their alignment based counterparts.

\subsection{Results on an Alignment Based Model}

To show  the versatility of DiMS, we conduct experiment on alignment-based
models leveraging Connectionist Temporal Classification
(CTC)~\citep{graves2006connectionist} objective.
Imputer~\citep{Saharia2020NonAutoregressiveMT} is among a few models that are
both alignment based and iterative. There is no official implementation of
Imputer available, therefore we implement a version ourselves (denoted with
$\dagger$) \footnote[1]{Based on the following implementation:
\url{https://github.com/rosinality/imputer-pytorch} }. 

Table~\ref{tab:imp} summarizes the results of DiMS applied to Imputer for both
directions of the WMT'14 English-German dataset. While DiMS boosts single step
translation of Imputer, it still falls behind more recent alignment based
models mentioned in Table \ref{tab:main}. However, we believe if one
incorporates various tricks introduced for alignment based models recently and
create a better iterative model, then DiMS can be an effective tool to further
enhance the single step translation. Details of Imputer training and
distillation are explained in Section~\ref{sec:sup-imp} of the appendix.

\begin{table}
{\footnotesize
\begin{tabular}{l c c}
    \toprule
    Method & {WMT'14 En-De} & {WMT'14 De-En}  \\
    \midrule
    Imputer$^{\small{\dagger}}$    &   25.9 & 29.0       \\
    Imputer$^{\small{\dagger}}$ + DiMS & \textbf{26.4} & \textbf{29.8}  \\
    \bottomrule
\end{tabular}
\captionof{table}{Single-step test set BLEU score for Imputer models trained on
  WMT'14 English-German.}
\label{tab:imp}
}
\end{table}

\subsection{DiMS on Raw Dataset}

The performance of the leading iNATs is at best similar to the autoregressive
model used for sequence level knowledge distillation. This limits the final
performance of iNATs and makes training without distillation
desirable~\citep{huang2021improving}. Table~\ref{tab-raw} shows that DiMS
improves the raw performance by a large margin even more than the corresponding
distilled variant. For instance, DiMS gets more than 12 BLEU scores
improvements on single-step evaluation of CMLMC. 

For one decoding pass, when raw variants of CMLMC are distilled with DiMS the
performance is superior to training on the distilled dataset (without DiMS).
This makes DiMS preferable to sequence-level knowledge distillation.
Nevertheless, the best performance is obtained when the two distillation
approaches are combined. 

\begin{table}[!h]
\centering
\begin{tabular}{l c c c}
\toprule

& \multicolumn{2}{c}{\textbf{CMLMC}}  \\
& Teacher  & Student  \\
\toprule
En-De  & 11.7 & 23.2 \\
De-En  & 16.4 & 29.3 \\
En-Ro  & 21.4 & 29.3 \\
Ro-En  & 21.8 & 32.7\\
\bottomrule
\end{tabular}
\captionof{table}{Comparison of student and teacher on raw dataset.}
\label{tab-raw}
\end{table}

\subsection{Unsupervised DiMS}

\label{sec:method-unsup}
In previous sections, we assume access to a parallel dataset and feed a
partially masked reference sentence to both student and teacher. One can use
the teacher to generate synthetic target sentences during the distillation.
This relaxes the dependence on the references and enables using monolingual
datasets for distillation. As usual, there is a trade-off between computation
and sample quality i.e. using more decoding passes leads to better data while
increasing the computational requirements.  

\begin{table}
\begin{tabular}{l c c}
    \toprule
    Method & {WMT'14 De-En} \\
    \midrule
    CMLM            & 22.77\\
    CMLM + U-DiMS     & 29.45 \\
    CMLM + DiMS           & \textbf{29.74}\\
    \midrule
    CMLMC            & 23.63\\
    CMLMC + U-DiMS    & 30.52\\
    CMLMC + DiMS          & \textbf{30.81}\\
    \bottomrule
\end{tabular}
\captionof{table}{Single-step test set BLEU score for models trained with U-DiMS.}
\label{tab:unsup}
\end{table}

We refer to this unsupervised
distillation variant as U-DiMS. Note that unsupervised only refers to the
distillation, and for training the teacher we still require access to a
parallel dataset. The only distinction between U-DiMS and DiMS is the usage of
synthetic data generated by the teacher and the remaining parts are untouched.
We run U-DiMS on WMT'14 De-En for CMLM and CMLMC using two iterative steps to
generate the synthetic samples. Table~\ref{tab:unsup} shows the effectiveness
of U-DiMS, obtaining a similar performance to DiMS.

\begin{table}[!h]
\begin{tabular}{l c}
    \toprule
    Method & $1$-Step BLEU \\ 
    \midrule
    CMLM  & 25.77 \\
    CMLM + DiMS  & 30.85 \\
    CMLM + DiMS - Hidden. & 28.69 \\
    \midrule
    CMLM + DiMS (T=$4$) & 31.04 \\
    CMLM + DiMS (T=$8$) & 30.97 \\
    \midrule
    CMLM + DiMS + EMA & \textbf{31.63} \\
    CMLM + DiMS (T=$4$) + EMA & 31.52 \\
    CMLM + DiMS (T=$8$) + EMA & 31.36 \\
    \bottomrule
\end{tabular}
\captionof{table}{BLEU score on WMT'16 En-Ro validation set with beam length
set to one as its done for early stopping. T stands for the number of teacher
decoding steps and is set to two if not specified.}
\label{table:ablation}
\end{table}

\subsection{Ablation Studies}
\vspace*{-0.1cm}

\begin{figure}[t]
    \includegraphics[width=\linewidth]{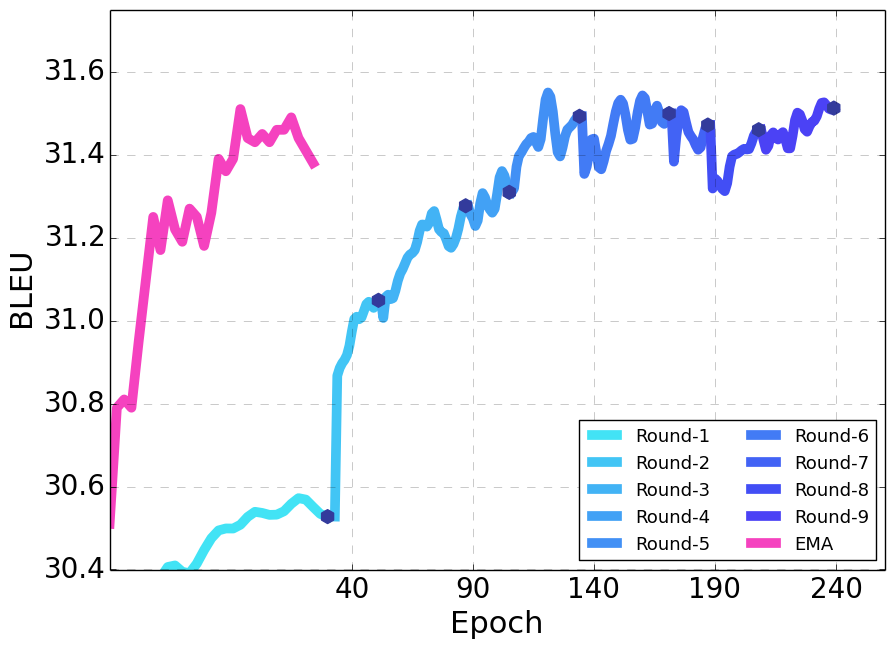}
    \captionof{figure}{Validation set BLEU score on WMT'16 En-Ro for iterative
    distillation and a EMA model. Each round of iterative distillation is shown
    with a unique color and the end of the round is noted by a black dot. The
    number of steps differs in various rounds as we use early stopping.}
    \label{fig:progressive}
\end{figure}

We conduct all the ablation studies on CMLM over WMT'16 En-Ro as it is
smaller than WMT'14 and validation set is used for evaluation.

\subsubsection{Hidden State Loss}
\label{sec:ablation-embed}
To investigate the effects of hidden state loss, we conduct an ablation study
in this section. The first block in Table~\ref{table:ablation} includes BLEU
scores for the base DiMS model with and without this term. The single-step
performance of the distilled model is improved over 2 BLEU points by leveraging
this loss. This supports the fact that the hidden states contain extra
information that is not available in soft labels. The exact value of $\lambda$
is selected based on a grid search reported in Section~\ref{sec:sup-abl} of the
appendix.

\subsubsection{EMA}

In order to establish the computational advantages of the slow-moving average,
we compare it with running the base variant for $9$ iterative rounds.
Figure~\ref{fig:progressive} demonstrates that the EMA variant is able to match
the iterative distillation with far fewer updates (almost equal to one round of
the distillation).

We observed that it is essential to move the teacher toward the student slowly.
For example, when $\mu \leq 0.9,$ the collapse to a degenerate solution
(explained in Section~\ref{sec:method-ema}) occurs before the end of the first
epoch. We plot the validation curve for various values of $\mu$ in
Section~\ref{sec:sup-ema} of the appendix showing the importance of the
slow-moving average.

\subsubsection{Teacher Decoding Steps} One hyper-parameter in DiMS algorithm is
the number of teacher's decoding steps. In order to investigate the effect of
this hyper-parameter, we set it to $2$, $4$, and $8$ while turning EMA on and
off. The two bottom blocks of Table~\ref{table:ablation} include the results of
this ablation. Although running the teacher for $4$ decoding steps shows
superior performance without EMA, as soon as we turn it on the gap disappears.
This shows that EMA can gradually improve the teacher and remove the need for
several iterative steps. Thus, we find no reason to set this hyper-parameter
larger than $2$ as it only increases distillation's computational cost.
\begin{figure}[t]
\centering
    \includegraphics[width=0.9\linewidth]{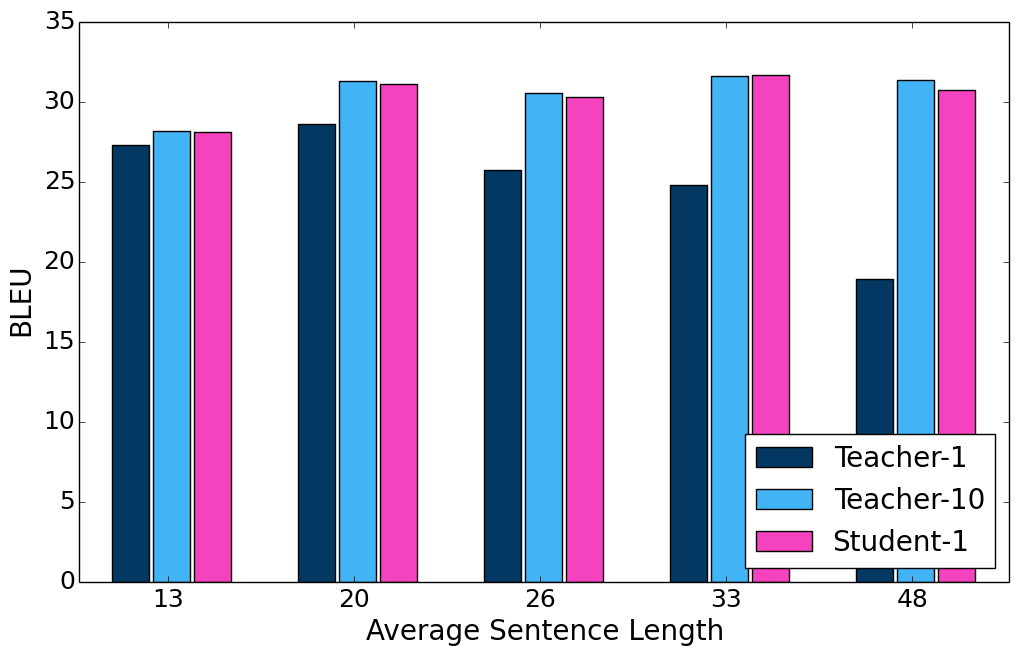}
    \captionof{figure}{Test set BLEU score on WMT'14 De-En based on the target
    sentence length for CMLM teacher and student.}
    \label{fig:length}
\end{figure}

\subsection{Analysis}

We study the effect of target sentence lengths on DiMS performance. The test
set is divided into five equally-sized buckets based on the target length. The
BLEU scores are reported for each bucket in Figure~\ref{fig:length}. The main
benefit of the iterative model is manifested by large sentences. The reason
might be the fact that longer sentences require a context and modeling it
becomes challenging with the conditional independence assumption in NAT. It is
clear in Figure~\ref{fig:length}, that the performance is improved in every
bucket. This improvement is most visible in the bucket with the highest average
sentence length. This is because of the fact that the same bucket has the
largest gap between the teacher's single and multi-step evaluation. 

\begin{figure}[t]
    \centering
    \begin{subfigure}[b]{0.25\textwidth}
        \centering
         \includegraphics[width=\textwidth]{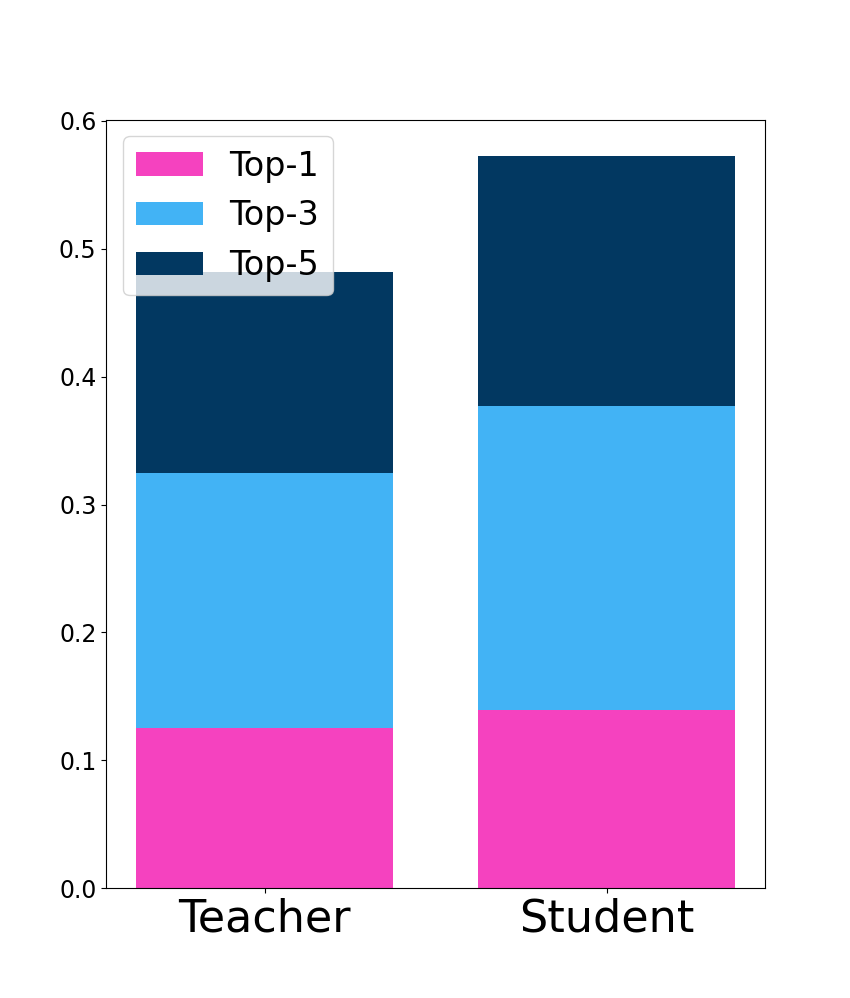}
         \caption{WMT'14 De-En} 
    \end{subfigure}
    \hskip -0.45cm
    \begin{subfigure}[b]{0.25\textwidth}
        \centering
        \includegraphics[width=\textwidth]{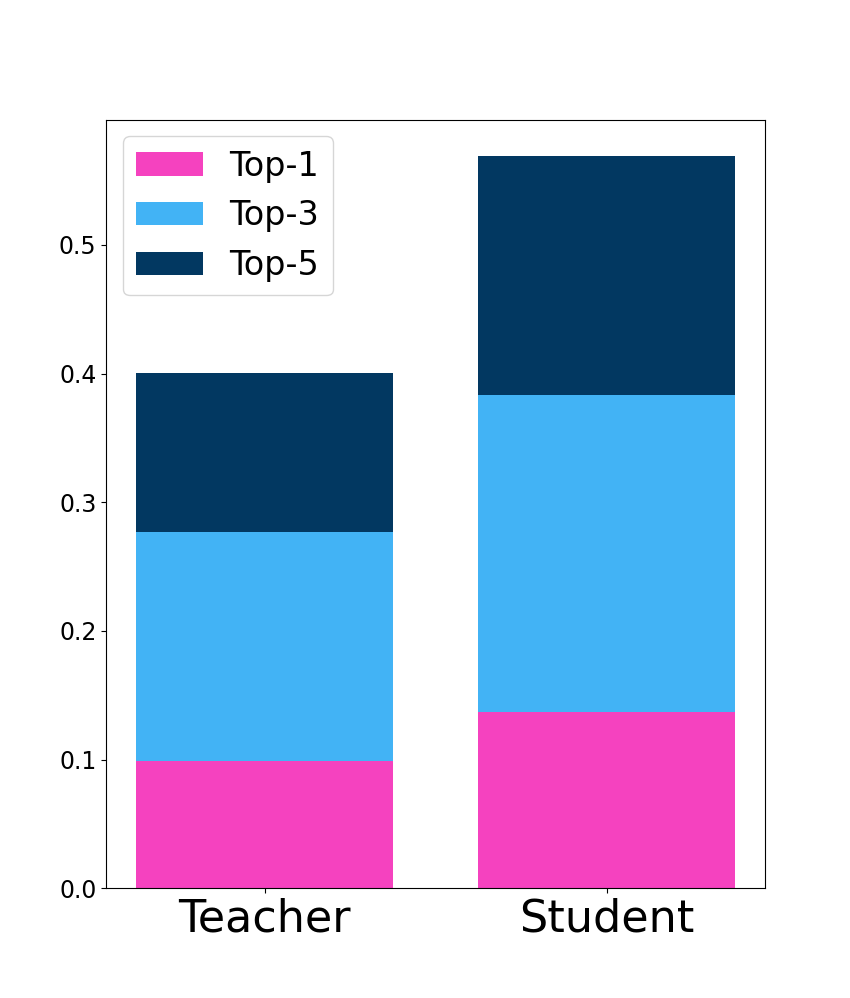}
        \caption{WMT'16 En-Ro}
    \end{subfigure}
    \caption{Comparison of CMLM teachers and students in
    predicting the target length. Top-1 means predicting the target correct
    and Top-3 and Top-5 means being incorrect by 1 and 2 offsets.}
    \label{fig:lp}
\end{figure}

We combine the length predictor objective with ours to account for changes in
the encoder's parameters. Interestingly enough, DiMS improves the performance
of the length predictor as depicted in Figure~\ref{fig:lp}. This shows that the
encoder benefits from the distillation as well.

Table~\ref{tab:qual} shows a qualitative example from the WMT'14 De-En dataset. The
improvements in samples are evident by comparing the predictions of the teacher
and the student with the target sentence. We provide more qualitative examples
in the appendix.

\section{Related Works}

\label{sec:related}
Many techniques have been proposed for iterative non-autoregressive machine
translation. Earlier attempts include denoising
autoencoder~\citep{lee2018deterministic} and
insertion-deletion~\citep{stern2019insertion,gu2019levenshtein}. More recently,
\citet{ghazvininejad2019mask} introduced the Mask-Predict improving the
performance of iNATs by employing a conditional masked language model.
CMLMC~\citep{huang2021improving} and SMART~\citep{ghazvininejad2020semi}
improve CMLM by incorporating a correction mechanism.
DisCo~\citep{kasai2020deep} is another variant conditioning each token on an
arbitrary subset of the other tokens. DiMS is entangled with the progress in
this domain as it requires a pre-trained iterative teacher.

The position constraint in cross-entropy can make the NAT training challenging,
therefore \citet{ghazvininejad2020aligned} propose aligned cross-entropy (AXE),
an objective that considers the best monotonic alignment between the target and
the model's predictions. \citet{du2021order} relaxes the monotonic assumption
and introduces Order Agnostic Cross-Entropy (OAXE).
CTC~\citep{libovicky2018end} is a similar alignment-based objective that fixes
the model output length and considers various alignments leading to the same
target. Imputer~\citep{Saharia2020NonAutoregressiveMT} extends CTC to benefit
from iterative refinements. 

GLAT~\citep{qian2021glancing} shows that the optimization challenges of iNATs
can be mitigated by introducing a curriculum learning focusing on sentences
with only a few masked tokens in the early stages of the training and gradually
increasing the masking ratio. ENGINE~\citep{tu2020engine} assumes access to a
pre-trained autoregressive model and optimizes a NAT model to maximize the
likelihood under the probability distribution defined by the pre-trained model.

\citet{salimans2021progressive} applies a distillation technique similar to
DiMS on generative models to decrease the number of required steps for
generating high-quality images. In contrast to DiMS, the distillation is
applied progressively. DiMS eliminates the need for progressive distillation by
updating the teacher with EMA. Lastly, the proposed EMA has some resemblance to
self-supervised learning techniques~\citep{grill2020bootstrap,
caron2021emerging, he2020momentum} where two models are updated, one through
gradient-based optimization and the other one through EMA. Despite this
similarity, the motivations are quite different. In self-supervised learning,
EMA is proposed as a technique to remove large negative sets whereas here EMA
enhances the quality of the labels generated by the teacher.

\begin{table*}[t]
\centering
\begin{tabular}{ l l}
\toprule
     Target & \multirow{2}{0.75\linewidth}{\footnotesize  The antibodies hunt down any nicotine molecules in the bloodstream , neutralising them before they reached the brain , preventing a smoker from getting a nicotine hit .}\\
     & \\
     \midrule
     Teacher &
     \multirow{2}{0.75\linewidth}{\footnotesize  The antibodies hunt the nicotine molecules molecblood neutralize them before reach brain a smoker not experience high nicotine .}\\
     & \\
     \midrule
     Student &
     \multirow{2}{0.75\linewidth}{\footnotesize The antibodies hunt the nicotine molecules in the blood and neutralize them before they reach the brain , so a smoker does not experience a nicotine high .}\\
     & \\
     \midrule
\end{tabular}
\captionof{table}{\small A qualitative example from WMT'14 De-En along with teacher and student's predictions on CMLMC.}
\label{tab:qual}
\end{table*}

\vspace*{0.2cm}
\section{Discussion}
\vspace*{0.2cm}
It is not completely clear why knowledge distillation works in
general~\cite{zhou2019understanding, huang2022learning}. But when it comes to
DiMS, we hypothesize that the labels generated by the teacher make the task
simpler for the student. In other words, it is difficult for the model to close
the gap between its single step prediction and ground truth while distillation
with teacher-generated labels reduces this gap. The importance of the gap
between labels and the model capacity has also been observed
before~\cite{mirzadeh2020improved}.

\section{Conclusion}

We introduce DiMS, an effective distillation algorithm that enhances the
single-step translation quality of a pre-trained iterative model. This is done
by replicating the model's multi-step behavior through one decoding pass. The
distillation can be repeated to achieve greater gains, but this increases the
training time noticeably. We show that the same benefits are obtainable by
setting the teacher as a moving average of the student while keeping the
training time comparable to one round of the distillation. Experiments over raw
and distilled datasets on four translation tasks for supervised and
unsupervised variants validate the effectiveness and versatility of DiMS.

Potential directions for future works include: \textbf{i)} The same family of
iterative models have been applied to automatic speech recognition, thus DiMS
is applicable to this domain. \textbf{ii)} One can combine a pyramid of
techniques introduced for iNATs to obtain a strong iterative model and make it
computationally efficient via DiMS. \textbf{iii)} Large monolingual sets can be
used to distill models with U-DiMS.

\section*{Limitations}
While DiMS makes the cross-entropy based family competitive with alignment
based variants, it still falls behind one some cases. Moreover, DiMS can
improve the performance of models trained on raw data, but the best performance
is still achieved when DiMS is applied on distilled datasets. Therefore, DiMS
still depends on an auto-regressive model for the best translation quality.

\section*{Acknowledgments}
We are grateful to Xiao Shi Huang for insightful comments and reviewing the
initial draft of this paper. We also thank Panteha Naderian for helpful
conversations.

\bibliography{custom}
\bibliographystyle{acl_natbib}

\appendix
\clearpage
\appendix

\begin{table*}[t]
\begin{center}
\begin{tabular}{l c c c c c}
    \toprule
    Method & Iteration & \multicolumn{2}{c}{WMT' 14} & \multicolumn{2}{c}{WMT' 16}  \\
     &  & En-De & De-En & En-Ro & Ro-En   \\
    \midrule

    CMLM\citep{ghazvininejad2019mask} & 10 & 27.0 & 30.5 & 33.1 & 33.3  \\
    CMLM & 10 & 26.9 & 31.2 & 33.1 & 33.6   \\
    \midrule
    CMLMC\citep{huang2021improving} & 10 & 28.4 & 31.4 & 34.6 & 34.1  \\
    CMLMC & 10 & 27.3 & 31.2 & 34.1 & 34.0 \\
    \midrule
    Imputer\citep{Saharia2020NonAutoregressiveMT} & 8 & 28.2 & 31.8 & 34.4 & 34.1  \\
    Imputer$^\dagger$ & 8 & 28.5 & 31.3 & - & - \\
    \bottomrule

    \bottomrule
\end{tabular}
\captionof{table}{\footnotesize Comparison of our teachers with the numbers
  reported in the original papers.}
\label{tab:teacher-comp}
\end{center}
\end{table*}

\section{Teacher Comparison}
\label{sec:sup-cmp}
Table~\ref{tab:teacher-comp} compares teachers trained by us with the 
original work proposing the model.

\section{EMA Momentum Effect}
\label{sec:sup-ema}
\begin{figure}[h]
    \centering
    \includegraphics[width=0.9\linewidth]{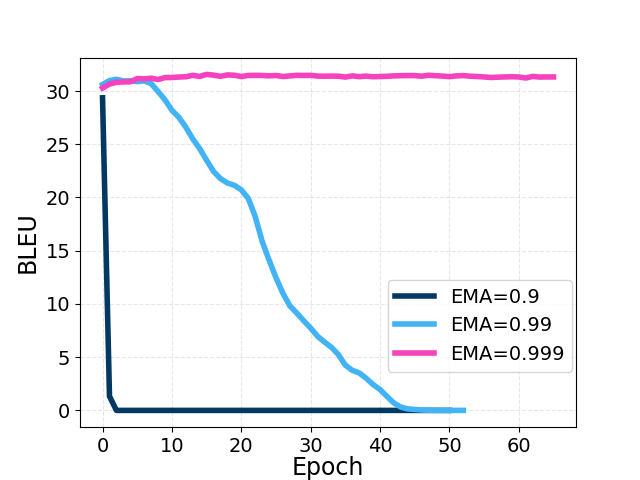}
    \caption{\footnotesize Validation BLEU curve on WMT'16 En-Ro for various
    EMA momenta with length beam=1.}
    \label{fig:my_label}
\end{figure}

We showcase the importance of the slow moving average in
Figure~\ref{fig:my_label}. As we increase the momentum the training becomes
more stable and leads to a better validation set BLEU score.

\section{Hyper-parameters for Distillation}
\label{sec:sup-hyp}
We use the same hyper-parameters for all the datasets.
{\small
\begin{center}
\begin{tabular}{l c c c}
    \toprule
    Hyper-parameter & CMLM/CMLMC \\
    \midrule
    Learning rate ($\eta$) & 1e-3\\
    Adam $\beta$ & (0.9, 0.98)\\
    Warm-up updates & 0 \\
    Max-tokens/GPU & 8192\\
    EMA momentum ($\mu$) & 0.9992\\
    Hidden state loss factor($\lambda$) & 0.7 \\
    Length loss factor & 0.1 \\
    Mask policy & Uniform\\
    Temperature & 0.5\\
    \bottomrule
\end{tabular}
\captionof{table}{Summary of hyper-parameters.}
\end{center}
}

\section{Ablation on Hidden State Loss Coefficient}
\label{sec:sup-abl}

The importance of the hidden state loss is shown in
Section~\ref{sec:ablation-embed} of the main body. We conduct an ablation study
in this section to find the optimal value of $\lambda$ that controls the
contribution of the hidden state loss.
\begin{figure}[h]
    \centering
    \includegraphics[width=0.9\linewidth]{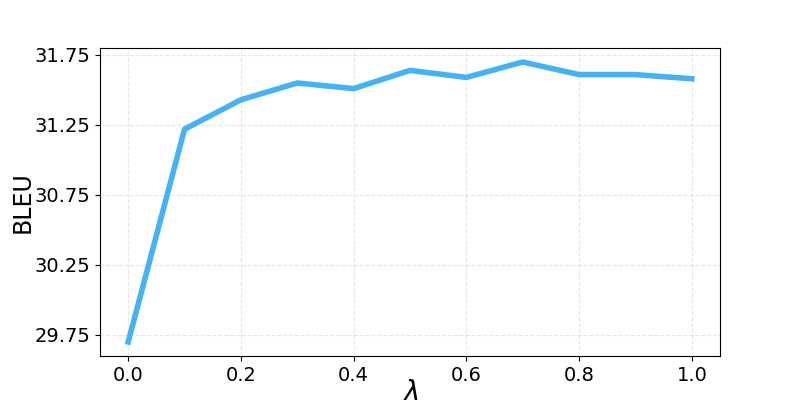}
    \caption{\footnotesize Best validation BLEU on WMT'16 En-Ro for CMLM with
    various hidden state loss coefficient ($\lambda$) }
    \label{fig:my_label2}
\end{figure}

\section{Computational Cost}
\label{sec:sup-comp}
\begin{figure}[h]
    \centering
    \includegraphics[width=0.6\linewidth]{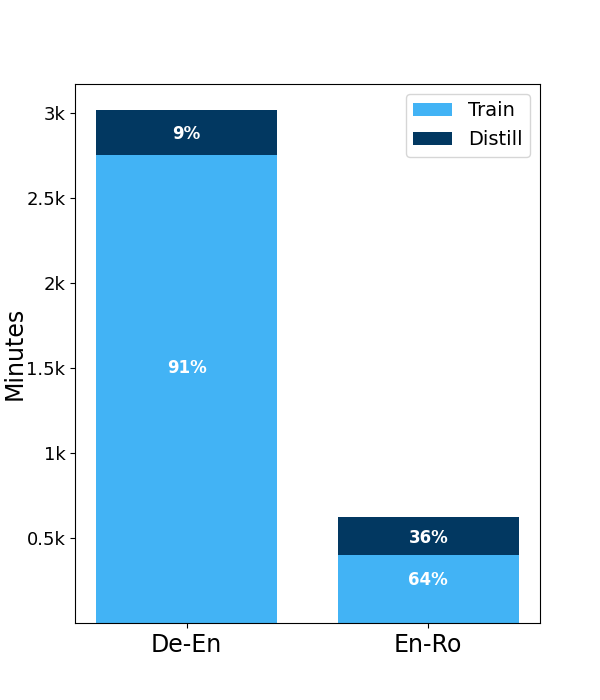}
    \caption{\footnotesize Comparison between teacher training and distillation time.}
    \label{fig:comp}
\end{figure}

During the distillation we have to run teacher for two steps which adds extra
computation. More specifically on a machine with 4 Nvidia-V100 32GB GPUs,
De-En training takes approximately 11 minutes per epoch compared to 27 minutes
for distillation and on En-Ro dataset training and distillation take 2 and 8
minutes per epoch, respectively. However, the number of epochs for distillation
is significantly less than teacher training. Precisely, teacher training takes
250 and 200 on De-En and En-Ro datasets respectively while  distillations takes
10 epochs for De-En and 30 epochs for En-Ro. Figure~\ref{fig:comp} compares the
overall time for training and distillation on De-En and En-Ro datasets and it
shows that the distillation time is one order of magnitude smaller than
training time.

Note that teacher is being run in the evaluation mode, thus the activations
maps are not kept in the memory. Therefore, the teacher can be run with a
larger batch-size which further reduces the computational costs. We leave this
as future works as it adds implementation complexity.

\section{Imputer Details}
\label{sec:sup-imp}
As mentioned in the main body, there is no official implementation of Imputer
available online. Here, we explain the differences between our implementation
and the original paper. Imputer proposes a pre-training phase where the model
is optimized merely with the CTC objective. We find it unnecessary as the model
reaches a better or competitive performance without it. Imputer leverages a
unified decoder rather than an encoder-decoder architecture incorporated here.
For Imputer training, computing the alignment with the highest probability is
necessary. This increases the training cost and
\cite{Saharia2020NonAutoregressiveMT} proposes either a pre-processing stage or
using a stale copy of the active model to manage the extra computation. We
compute the best alignment on the fly as it is still computationally feasible.
Similar to Imputer inference, extra care is taken to make sure consecutive
tokens are not unmasked in the same step. Instead of a Bernoulli masking policy
during training, we used a block masking policy.

For the distillation, Imputer mainly benefits from two
iterative steps and the gains are not as significant after that. Therefore,
there is no incentive to use EMA.

\label{sec:sup-qual}
\begin{table*}
\begin{tabular}{ l l}
    \multicolumn{2}{l}{\textbf{CMLM ~ WMT'14 De-En}} \\
    \midrule
     Target & \multirow{3}{0.75\linewidth}{\footnotesize The rate of 3.1 per cent is indeed better than the previous year and is also better than in September , " however , we had hoped for more , " said Monika Felder - Bauer , acting branch manager of the Employment Agency in Sonthofen .}\\
     & \\
     & \\
     \midrule

     Teacher & 
     \multirow{3}{0.75\linewidth}{\footnotesize  Although the quota was better 3.1 better than last year and better than September , we would hoped more , " Monika - Bauer Deputy of the Labour Agency in Sonthofen .}\\
     & \\
     & \\
     \midrule
     Student &
     \multirow{3}{0.75\linewidth}{\footnotesize Although the quota at 3.1 $\%$ is better than last year and is also better than in September , " we would have hoped for more , " says Monika Felder - Bauer , deputy head of the Labour Agency in Sonthofen .}\\
     & \\
     & \\
    \multicolumn{2}{l}{\multirow[b]{2}{\linewidth}{\textbf{CMLM ~ WMT'16 Ro-En}}} \\
    & \\
    \midrule
     Target & \multirow{3}{0.75\linewidth}{\footnotesize we must ask these people to learn the language , try to appropriate our values , to stop having one foot in europe and one in their home country , bringing the rest of the family including through marriages of convenience .}\\
     & \\
     & \\
     \midrule

     Teacher & 
     \multirow{3}{0.75\linewidth}{\footnotesize  let us ask these people to learn their language , try to take values , stop longer stand in europe and with one their country home origin , bringing the rest of family , including through convenience marriages .}\\
     & \\
     & \\
     \midrule
     Student &  
     \multirow{3}{0.75\linewidth}{\footnotesize let us ask these people to learn the language , try to take over values , no longer stand in europe and with one in their home country , bringing the rest of the family , including through convenience marriages .}\\
     & \\
     & \\
     \multicolumn{2}{l}{\multirow[b]{2}{\linewidth}{\textbf{CMLMC ~ WMT'14 De-En}}} \\
     & \\
     \midrule
     Target & \multirow{3}{0.8\linewidth}{\footnotesize Edward Snowden , the US intelligence whistleblower , has declared that he is willing to travel to Berlin to give evidence to the German parliament if the US National Security Agency and its director Keith Alexander fail to provide answers about its activities .}\\
     & \\
     & \\
     \midrule

     Teacher & 
     \multirow{3}{0.8\linewidth}{\footnotesize  Edward Snowden , the whistleblower of the US intelligence , has that he is to travel to Berlin and testify the German destag if the National Security Agency its director Keith Alexander not provide answers about their activities .}\\
     & \\
     & \\
     \midrule
     Student &  
     \multirow{3}{0.8\linewidth}{\footnotesize Edward Snowden , the whistleblower of the US intelligence , has said that he is prepared to travel to Berlin and testify to the German destag if the American National Security Agency and its director Keith Alexander do not provide answers to their activities .}\\
     & \\
     & \\

\end{tabular}
\end{table*}

\begin{table*}[!t]
\begin{tabular}{ l l}
    \multicolumn{2}{l}{\multirow[b]{2}{\linewidth}{\textbf{CMLMC ~ WMT'16 Ro-En}}} \\
    & \\
    \midrule
     Target & \multirow{3}{0.75\linewidth}{\footnotesize during the routine control , the border policemen observed in the truck 's cab , a large travel bag , which contained personal things of many people , which is why they conducted a thorough check on the means of transport .}\\
     & \\
     & \\
     \midrule

     Teacher & 
     \multirow{3}{0.75\linewidth}{\footnotesize  at specific control , border police officers a large travel of travel the cabvan , where there were things for several people , which is why carried out thorough control over the vehicle of transport .}\\
     & \\
     & \\
     \midrule
     Student &  
     \multirow{3}{0.75\linewidth}{\footnotesize at specific control , border police observed , in the cabin , a large travel getravel , where there were personal things for several people , which is why they carried out thorough control over the vehicle of transport .}\\
     & \\
     & \\
\end{tabular}
\end{table*}

\end{document}